\pdfoutput=1

\documentclass[11pt]{article}

\usepackage{acl}

\usepackage{times}
\usepackage{latexsym}

\usepackage[T1]{fontenc}

\usepackage[utf8]{inputenc}

\usepackage{microtype}
\usepackage{amssymb}
\usepackage{amsopn}
\usepackage{graphicx}
\usepackage{multirow}
\usepackage{makecell}
\usepackage[english]{babel}
\usepackage{bm}
\usepackage{array}
\usepackage{setspace}
\usepackage{CJKutf8}
\usepackage{algorithmicx}  
\usepackage{algpseudocode}  
\usepackage{amsmath}  
\usepackage{color,xcolor}
\usepackage{algorithm} 

\usepackage{pifont}
\usepackage{bm}
\usepackage{amsfonts}

\usepackage{graphicx}
\usepackage[normalem]{ulem}
\usepackage{multirow}
\usepackage{amsmath}
\usepackage{url}

\usepackage{tikz}
\usepackage{xcolor}
\usepackage{tcolorbox}
\usepackage{color,xcolor}

\usepackage{amssymb}
\usepackage{amsopn}
\usepackage{graphicx}
\usepackage{multirow}
\usepackage[english]{babel}
\usepackage{bm}
\usepackage{array}
\usepackage{setspace}
\usepackage{todonotes}

\usepackage{graphicx}
\usepackage{CJKutf8}
\usepackage{xspace}
\usepackage{amsfonts}
\usepackage{array,multirow}
\usepackage{pgfplots}
\usepackage{tikz}
\usepackage{verbatim}
\usepackage{arydshln}
\usepackage{booktabs}
\definecolor{ugreen}{rgb}{0,0.5,0}
\definecolor{mygreen}{RGB}{58,127,88}
\definecolor{iyellow}{RGB}{255,250,205}
\definecolor{ipurple}{RGB}{230,230,250}
\definecolor{myred}{RGB}{160,52,52} 
\definecolor{myblue}{RGB}{30,144,255}
\definecolor{myorange}{RGB}{255,127,80}
\definecolor{mypurple}{RGB}{255,20,147}

\usepackage{mathtools}
\DeclarePairedDelimiterX\set[1]\lbrace\rbrace{#1}

\newcommand{\cbleu}[1]{\textcolor{blue}{#1}}

\title{LexMatcher: Dictionary-centric Data Curation \\ for LLM-based Machine Translation}

\author{Yongjing Yin$^{1,2}$, Jiali Zeng$^{4}$, Yafu Li$^{2}$, Fandong Meng$^{4}$, Yue Zhang$^{2,3}\thanks{Corresponding author}$ \\
$^{1}$Zhejiang University \\
$^{2}$School of Engineering, Westlake University \\
$^{3}$Institute of Advanced Technology, Westlake Institute for Advanced Study\\
$^{4}$Pattern Recognition Center, WeChat AI, Tencent Inc \\
{\tt \{yinyongjing,liyafu\}@westlake.edu.cn} \\ 
{\tt \{lemonzeng,fandongmeng\}@tencent.com} \\
{\tt yue.zhang@wias.org.cn }
}

\begin{document}
\maketitle
\begin{CJK}{UTF8}{gbsn}

\begin{abstract}

The fine-tuning of open-source large language models (LLMs) for machine translation has recently received considerable attention, marking a shift towards data-centric research from traditional neural machine translation. 
However, the area of data collection for instruction fine-tuning in machine translation remains relatively underexplored. 
In this paper, we present LexMatcher, a simple yet effective method for data curation,
the design of which is driven by the coverage of senses found in bilingual dictionaries. 
The construction process comprises data retrieval from an existing corpus and data augmentation that supplements the infrequent senses of polysemous words. 
Utilizing LLaMA2 as our base model, our approach outperforms the established baselines on the WMT2022 test sets and also exhibits remarkable performance in tasks related to word sense disambiguation and specialized terminology translation. 
These results underscore the effectiveness of LexMatcher in enhancing LLM-based machine translation.

\end{abstract}

\section{Introduction}

The emergence of large language models (LLMs) \cite{gpt3,llama2,arxiv2023:openai_gpt4} has brought about new opportunities for machine translation
and improving the translation performance of smaller-sized LLMs (7B or 13B) has attracted a lot of attention \cite{jiao-etal-2023-parrot,zeng-etal-2023-tim,Arxiv2023:bayling,alma}.
Unlike traditional neural machine translation (NMT)
which relies heavily on abundant parallel data \cite{DBLP:conf/acl/SennrichHB16,DBLP:conf/emnlp/EdunovOAG18,gordon-etal-2021-data,DBLP:conf/icml/FernandesGGFF23}.
LLMs have demonstrated less dependency on vast amounts of supervised data to achieve competitive performance.
Similar to other tasks by LLMs \cite{lima,textbook}, the quality of fine-tuning data plays a more crucial role in NMT \cite{Arxiv2023:bayling,alma}.

Current work primarily focuses on constructing fine-tuning data by leveraging human-written development sets, and creating refined instruction data for special purposes such as contrastive translation pairs and interactive translation \cite{zeng-etal-2023-tim,Arxiv2023:bayling}. 
However, these methods do not fully exploit the potentially valuable information embedded within the existing large parallel corpus.
It has been demonstrated that fine-tuning LLMs with extensive parallel data can impair their inherent translation capabilities \cite{alma}. 
Moreover, the quality of data distributions has been emphasized to have a more significant impact on the model performance than quantity alone \cite{textbook,textbook2}, with more uniform data distributions contributing to improved generalization for unseen compositions \cite{patel-etal-2022-revisiting}.

Motivated by the above observations, we investigate a principled method, LexMatcher, for curating supervised fine-tuning data for LLM-based translation.
The objective is to collect a small yet carefully selected dataset that follows a proper distribution for maximizing translation quality. 
To this end, we leverage a bilingual dictionary as a pivotal resource to ensure comprehensive coverage of word or phrase senses in bilingual contexts.
The construction of the dataset involves two steps: data retrieval and data augmentation. 
In the data retrieval step, we traverse commonly-used corpora (e.g., WMT training data) and \textbf{identify} sentence pairs that are guided by the coverage of dictionary senses.
Inevitably, however, there may be uncovered senses of polysemous words, 
representing long-tail knowledge essential for accurate translation. 
In the data augmentation step, we employ a commercial LLM (e.g., ChatGPT) to \textbf{generate} precise and concise sentence pairs that contain the uncovered senses.
Finally, we fine-tune LLMs using a combination of the retrieved and generated data.

We conduct extensive experiments on six language directions including Zh$\Leftrightarrow$En, En$\Leftrightarrow$De, and En$\Leftrightarrow$Ru. 
By employing LexMatcher, we extract 0.1\% of the WMT data, totaling
1 million samples across all six language directions.
Results of fine-tuned LLMs on the test sets show the superiority of our method over the baselines in both standard and zero-shot settings. 
The fine-tuned models also achieve comparable or better performance in terminology translation and translation disambiguation compared to the dedicated or commercial systems. 
Further analyses of different data collection methods and composition generalization underscore the significance of high-quality data distributions.
We will release the code, data, and models upon acceptance.

\section{Related Work}

\paragraph{Data Selection for NMT.}

For traditional neural machine translation models, augmenting the volume of parallel data often leads to improvements in performance \cite{DBLP:conf/acl/SennrichHB16,DBLP:conf/emnlp/EdunovOAG18,gordon-etal-2021-data,DBLP:conf/icml/FernandesGGFF23}.
Conversely, there have also been studies exploring data selection to reduce the size of the training corpus.
For instance, \citet{van-der-wees-etal-2017-dynamic} gradually reduces the training data to a cleaner subset, determined by external scorers. 
\citet{wang-etal-2018-denoising} introduce curriculum-based data selection that employs a trusted clean dataset to assess the noise level of each sample. 
\citet{kumar-etal-2019-reinforcement} employ reinforcement learning to simultaneously learn a denoising curriculum and improve the NMT model. 
\citet{mohiuddin-etal-2022-data} initially train a base NMT model on the entire available data and subsequently fine-tune the base model using selected subsets.
Compared to traditional NMT, data curation is more critical for LLM-based MT, for which we make the first investigation by proposing a simple and practical method.

\paragraph{LLMs for MT.}
The usage of LLM-based MT is significantly different from the conventional NMT. 
LLMs, particularly large ones like GPT-4, serve as interfaces that can perform translation with simple translation instructions or in-context learning (ICL) \cite{emnlp/LinMAWCSOGBDPSK22,Hendy,Zhu,Agrawal}. 
For ICL, the influence of data selection methods on model performance is not significantly noticeable \cite{Zhu,Agrawal,emnlp/LinMAWCSOGBDPSK22}.
Fine-tuning smaller-sized LLMs such as LLaMA \cite{arxiv2023:LLaMA} for translation has garnered increasing attention \cite{jiao-etal-2023-parrot,Arxiv2023:bayling},
which has the potential to achieve an improved trade-off between quality and efficiency. 
TIM \cite{zeng-etal-2023-tim} constructs translation pairs for comparison and introduces an additional preference loss. 
Bayling \cite{Arxiv2023:bayling} automatically generates interactive translation instructions. 
\citet{Mao:constrastive} construct an additional cross-lingual discrimination task using word alignment for low-resource languages.
\citet{BigTrans} fine-tune LLMs using more than 300 million parallel instances while \citet{alma} indicate that such strategy could potentially impair the translation capabilities of LLMs. 
Instead, they propose a two-stage process that involves further post-training LLMs using a substantial amount of mixed monolingual data, followed by a subsequent step of fine-tuning with human-written parallel data.

In line with the above efforts, we also aim to improve the open-source LLMs.
The difference is that we propose specific parallel data collection methods, following the principle of achieving uniform coverage of semantic units in the dictionary. 
Moreover, our approach achieves a better balance between efficiency and performance, and we can obtain a high-quality translation model using fewer computational resources compared to continual pretraining.

\paragraph{Bilingual Dictionary for NMT.}
Bilingual dictionaries have been employed to enhance translation quality, particularly for rare words or domain-specific entities. 
One approach involves augmenting the training data with pseudo-parallel sentences generated based on the dictionary.
For example, \citet{DBLP:conf/ijcai/ZhaoZZZ20} enhance the parallel corpus with the help of paired entities extracted from multilingual knowledge graphs.
\citet{hu-etal-2022-deep} propose denoising entity pretraining for NMT using monolingual data and paired entities.
These methods do not consult bilingual dictionaries for translation candidates during the inference stage.
Another approach involves leveraging bilingual alignments as lexical constraints \cite{li-etal-2022-prompt,wang-etal-2022-integrating,zeng-etal-2023-extract}.
For LLMs, bilingual dictionaries have been used as a part of prompts \cite{cod_llm,Dic_prompt_llm} for the LLMs of more than 100B.
In contrast, we aim to improve LLMs' fine-tuning performance on translation tasks. The dictionaries serve as a pivot for data collection and can also be added in prompts when needed.

\begin{figure}[t]
\centering
\includegraphics[width=1.0\linewidth]{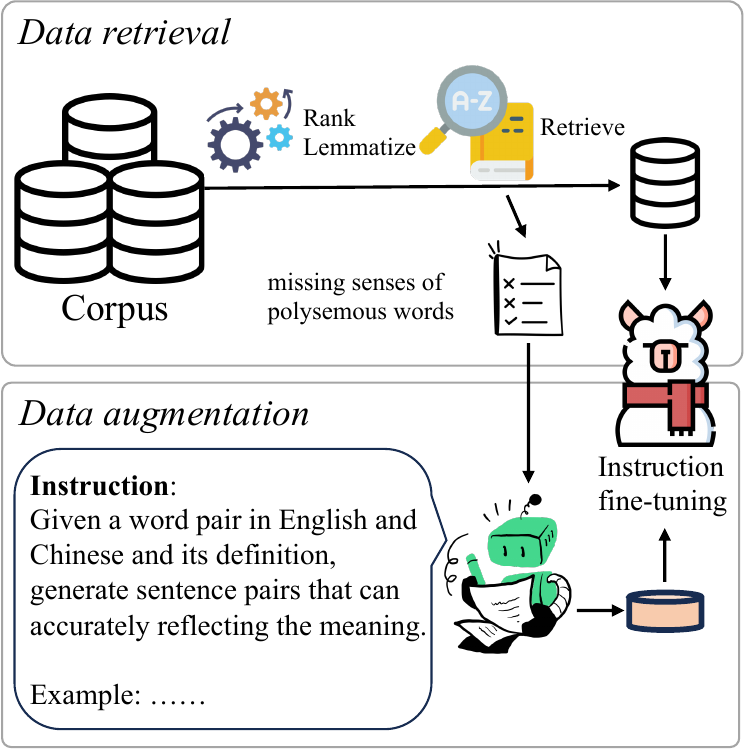}
\caption{
Illustration of our LexMatcher for instruction fine-tuning smaller LLMs (e.g., LLaMA).
}
\label{fig:method}
\end{figure}

\section{Method}
\label{sec_method}

The overview of LexMatcher is illustrated in Figure \ref{fig:method}, which takes data retrieval (\S\ref{sec_data_retrieval}) and data augmentation (\S\ref{sec_data_augmentation}) steps for curating a compact parallel dataset for instruction fine-tuning.

\subsection{Data Retrieval}
\label{sec_data_retrieval}

Given a dictionary $\Phi={(s,t)}$, where $\Phi=\{(s_1, t_1), (s_2, t_2), \ldots, (s_n, t_n)\}$ and each $(s_i, t_i)$ represents a source-target segment pair, we aim to ground each pair in parallel contexts by retrieving data from a bilingual parallel dataset $D=\{(x,y)\}$. 
The dictionary $\Phi$ shares the same source and target languages with $D$.
The segments can be words (e.g., ``country''), phrases (e.g., ``take over''), or named entities (e.g., ``World Trade Organization'') in the dictionary.
Ideally, the objective is to find a subset $S_r \subseteq D$ such that:
\begin{equation}
    \forall (s, t) \in \Phi, \exists (x, y) \in S_r : s \subseteq x \wedge t \subseteq y,
\end{equation}
where $x=\{x_1,x_2,...,x_{|x|}\}$ and $y=\{y_1,y_2,...,y_{|y|}\}$.
In practice, however, 
it is not guaranteed that the existing bilingual corpora can cover all dictionary senses. Therefore, we extract a subset that strives to fulfill this objective.

We traverse the corpus in sequential order and search for potential matches with segment pairs in the dictionary.
To prioritize the extraction of high-quality sentence pairs, we rank the corpus with model-based translation quality metrics, e.g., COMET-KIWI \citep{rei-etal-2022-comet22}. 
Specifically, for each segment\footnote{We use unigram and bigram excluding stopwords.} in a source sentence, we perform a dictionary lookup for all the aligned target words.
If one of the aligned target segments exists in the target sentence, we put the sentence pair into the translation candidate subset $S_r$.
We lemmatize each word in the source and target sentence to alleviate the effect of morphological textual variations. 
In addition, we introduce a threshold $K$ to skip the sentence if all the segment pairs in it have already been matched $K$ times.
$K$ enables convenient control over the size of the subset and is used to encourage even distribution of segment pairs. 
The matching procedure is illustrated in Algorithm~\ref{alg:main}.

\begin{algorithm}[t!]
\caption{Data retrieval in LexMatcher}
\begin{algorithmic}[1]
\State \textbf{Input:} Parallel dataset $D$, dictionary $\Phi$, threshold $K$
\State \textbf{Output:} Subset $S_r \subseteq D$
\State Initialize $S_r\leftarrow\emptyset$，frequency count $C\leftarrow\{\}$
\For{each $(x, y) \in D$}
    \State \textit{Found}$\leftarrow$false
    \For{each segment $\hat{x_i}$ in Lemmatize($x$)}
        \For{each $t_n$ in $\Phi[\hat{x_i}]$}
            \If{$C[(\hat{x_i},t_n)]<K$ \textbf{and} 
\\ \ \ \ \ \ \ \ \ \ \ \ \ \ \ \ \ \ \ \ \ \ \ $t_n$ in Lemmatize($y$)}
                \State $C[(\hat{x_i},t_n)]$$\leftarrow$$C[(\hat{x_i},t_n)]+1$
                \State \textit{Found}$\leftarrow$true
            \EndIf
        \EndFor
    \EndFor
    \If{\textit{Found}}
        \State Add $(x, y)$ to $S_r$
    \EndIf
\EndFor
\State \textbf{return} $S_r$
\end{algorithmic}
\label{alg:main}
\end{algorithm}


\subsection{Data Augmentation}
\label{sec_data_augmentation}
Using a partial set of open-source corpora cannot cover all the senses in the dictionary, which can be rare named entities or low-frequency occurrence of distinctive senses of certain words.
The translation of rare entities is generally unique and can be solved effectively by prompting LLMs during inference, and the lack of training data for these cases may have minimal impact. 
However, the senses of polysemous words are context-sensitive and may require specific training data to strengthen the model's understanding and translation of them.
To compensate for the missing senses, we leverage ChatGPT\footnote{GPT-3.5-turbo-0314} to construct translation demonstrations for each sense, thus creating the subset $S_c$.
Concretely, we prompt ChatGPT with a sense expressed in source and target languages and the sense's definition.
The prompt is shown in Figure~\ref{fig:prompt_gpt} (Appendix \ref{app_chatgpt}).
Only nouns and verbs with more than three senses are considered due to their highly polysemous nature \cite{campolungo-etal-2022-dibimt}.
Note that the subset $S_c$ only takes up a neglectable portion of the whole dataset (e.g., 225 sentence pairs for English-Germen, and the specific numbers are reported in \S\ref{sec_exp}).


\section{Instruction Fine-tuning LLM for MT}

Instruction fine-tuning has become standard practice in LLM-based translation \cite{zeng-etal-2023-tim,alma,Arxiv2023:bayling}. 
Our instruction-following data is constructed based on $S=S_r\cup S_c$ (\S\ref{sec_method}).
Generally, each instance comprises an ``instruction'' $c$ describing the task the model should perform (e.g., ``Translate the sentences from English to Chinese.''), an ``input'' $x$ indicating the source sentence, and a corresponding output $y$ indicating the answer to the instruction, i.e., the target sentence.
The LLMs are optimized by minimizing the negative log-likelihood of the output $y$:
\begin{equation}
    L=-\sum_{(x,y)\in S}\frac{1}{|y|}\sum_i^{|y|}\log p(y_i|c,x;\theta),
\end{equation}
where $\theta$ is the trainable parameters.

We use two kinds of translation instructions: 1) general translation instructions mainly used to indicate translation directions (e.g., ``Translate the following sentence to English''),
and 2) constrained translation instructions that specify word translations from a given dictionary or based on specific user requirements.
(e.g., `Translate the following sentence to English using the given reference translations.'')
For the latter, we randomly sample a small number of sentence pairs to incorporate specified word translations\footnote{The maximum number of sentences under the constrained translation instructions for each direction is set to 10,000.}.
For each sample, we introduce at most 3 segment pairs matched in the dictionary and orgnize them with a template:
\begin{equation}
    c = \text{Template}(\{(s_i, t_i)\}_{i=1}^{N}),
\end{equation}
where $s_i$ and $t_i$ denote the segment pair following Section \ref{sec_method}.
We simply use ``means’’ to connect $s_i$ and $t_i$, and prepend the constraint to the translation instruction.
An example is shown in Figure \ref{fig:prompt_gpt}(b) (Appendix \ref{app_chatgpt}).
During inference, we can choose whether to use the constrained translation instructions to incorporate translations from the dictionary or terminology, depending on the situation.

\section{Experiments}
\label{sec_exp}



\subsection{Setting}

\begin{table}[t]
\centering
\small
\begin{tabular}{lccccc}
\toprule
\multirow{2}{*}{Lang} & \multirow{2}{*}{Raw} & \multicolumn{3}{c}{Retrieval} & \multirow{2}{*}{Supplement} \\
&    & K=1  & K=2 & K=3 &  \\
\midrule
Zh & 33M & 75k &  188k & 281k & 2.2k  \\
De & 278M & 93k &  233k & 351k & 0.2k \\
Ru & 227M & 98k &  246k  & 367k & 0.7k \\
\bottomrule
\end{tabular}
\caption{
The number of parallel sentences of different data sets.
}
\label{data_statis}
\end{table}

\begin{table*}[!t]
\centering
\small
\renewcommand{\arraystretch}{1.1}
\begin{spacing}{1.1}
\resizebox{\textwidth}{!}{
\begin{tabular}{lcccccc}
\toprule
\multirow{2}{*}{\bf Model} &
 \multicolumn{1}{c}{\bf Zh$\Rightarrow$En} & \multicolumn{1}{c}{\bf En$\Rightarrow$Zh} & \multicolumn{1}{c}{\bf De$\Rightarrow$En} & \multicolumn{1}{c}{\bf En$\Rightarrow$De} &
 \multicolumn{1}{c}{\bf Ru$\Rightarrow$En} & \multicolumn{1}{c}{\bf En$\Rightarrow$Ru} \\
& {BLEU}/{COMET} & {BLEU}/{COMET} & {BLEU}/{COMET} & {BLEU}/{COMET} & {BLEU}/{COMET} & {BLEU}/{COMET} \\
\midrule
$\text{GPT-3.5}^{\dag}$ & 26.60/82.90 & 44.90/87.00 &  33.10/85.50 & 34.40/87.00 & 42.40/86.10 & 34.40/87.00 \\
$\text{GPT-4}^{\dag}$ & 27.20/82.79 & 43.98/87.49 & 33.87/85.62 & 35.38/87.44 & 43.51/86.18 & 30.45/88.87 \\
$\text{NLLB-54B}^{\dag}$ & 16.56/70.70 & 27.38/78.91 &  26.89/78.94 & 34.50/86.45 & 26.89/78.94 & 30.96/87.92 \\
\midrule
$\text{LLaMA2-7B}^{\dag}$ & 18.19/75.00 & 16.97/71.80 & 30.42/82.74 & 19.00/76.39 & 36.02/82.84 & 16.00/73.24 \\
Parrot-7B \cite{jiao-etal-2023-parrot} & 20.20/75.90 & 30.30/80.30 & 27.30/82.40 & 26.10/81.60 & - & - \\
TIM-7B \cite{zeng-etal-2023-tim} & 24.51/79.71 & 37.83/85.10 & 26.12/78.94 & 20.90/74.91 & - & - \\
ALMA-7B \cite{alma} & 23.52/\textbf{79.73} & 36.48/85.05 &  29.49/83.98 & 30.31/85.59 & 38.93/\textbf{84.81} & 27.09/87.17 \\
LexMatcher-7B & \textbf{24.81}/79.13 & \textbf{40.34}/\textbf{86.11}	& \textbf{32.33}/\textbf{84.29} & \textbf{33.56}/\textbf{86.31} & \textbf{41.01}/84.43 & \textbf{28.97}/\textbf{87.23} \\
\midrule
$\text{LLaMA2-13B}^{\dag}$ & 21.81/78.10 & 30.00/79.70 & 31.06/83.01 & 13.69/75.55 &  36.50/82.91 & 0.59/63.84 \\
DictPrompt-13B \cite{Dic_prompt_llm} & 17.55/74.12 & 33.75/83.46 & 30.36/83.31 & 25.24/80.89 & 37.70/81.95 & 21.98/81.00 \\
BigTrans-13B \cite{BigTrans} & 14.16/74.26 & 28.56/81.31 & 23.35/80.68 & 21.48/78.81 & 26.81/77.80 & 17.66/78.21 \\
Bayling-13B \cite{Arxiv2023:bayling} & 20.12/77.72 & 37.92/84.62 & 27.34/83.02 & 25.62/82.69 & 33.95/82.07 & 12.77/71.01 \\
ALMA-13B \cite{alma} & 25.46/{\bf 80.21} & 39.84/85.96  & 31.14/{\bf 84.56} & 31.47/85.62 & 40.27/{\bf 85.27} & 28.96/87.53 \\
LexMatcher-13B & {\bf 26.15}/79.88	& {\bf 41.13}/{\bf 86.58} & {\bf 32.59}/84.55 & {\bf 34.82}/{\bf 86.45} & {\bf 41.53}/84.91 & {\bf 30.20}/{\bf 87.83} \\
\bottomrule
\end{tabular}
}
\end{spacing}
\caption{
\label{tab_results_main_result}
Evaluation results on WMT22 test sets. Higher scores (BLEU and COMET) denote better translation performance.
Bold numbers indicate the best scores among models of the same sizes.
The numbers with the dagger symbol represent the results from \cite{alma}.
LexMatcher-7B outperforms Parrot-7B and ALMA-7B with p-value<0.01, and LexMatcher-13B outperforms ALMA-13B with p-value<0.01.
}
\end{table*}

For parallel training data, we use the open-source data from WMT22\footnote{https://www.statmt.org/wmt22/translation-task.html} in German$\Leftrightarrow$English, Chinese$\Leftrightarrow$English, and Russian$\Leftrightarrow$English.
The detail of data preprocessing is shown in Appendix \ref{app_filter}.
We use bilingual dictionaries provided by Open Multilingual WordNet \cite{bond-etal-2016-cili}\footnote{https://www.nltk.org/howto/wordnet.html}.
In addition, we take Wikititles\footnote{https://data.statmt.org/wikititles/v3/} as an entity dictionary.
Table \ref{data_statis} presents the number of sentence pairs for each language pair in different subsets, including the original training set, subsets extracted based on different $K$, and the ChatGPT-generated data. 
It can be observed that our method achieves a high compression rate.
The subset $K$=3 is used for the main experiment, and the extracted data for Chinese, German, and Russian accounts for only 0.57\%, 0.08\%, and 0.11\% of the original data, respectively.
The development sets from the previous WMT competitions are used by default \cite{jiao-etal-2023-parrot,alma}.

We use LLaMA2-7B and LLaMA2-13B for comparing to the related methods, and one model is used for all of the translation directions.
We fine tune all of our models for 1 epoch with the collected multilingual instruction data.
The batch size is 128 and the learning rate is 2e-5. 
The final checkpoint is used for evaluation, and we use beam search with a beam size of 4 during inference.
For automatic evaluations, we use BLEU \cite{papineni2002bleu:} \footnote{https://github.com/mjpost/sacrebleu} and COMET\footnote{https://huggingface.co/Unbabel/wmt22-comet-da}.


\subsection{Main Results}

\paragraph{Seen Language Directions.}


Table \ref{tab_results_main_result} presents the translation performance on the WMT22 test sets.
The LLaMA2 models fine-tuned on the instruction data collected by LexMatcher significantly outperform their original zero-shot performance, especially for the En$\Rightarrow$xx.
Concretely, LexMatcher-7B improves LLaMA2-7B by an average of 17.02 BLEU points and 12.68 COMET points in En$\Rightarrow$xx, and by 4.45 BLEU points and 2.42 COMET points in xx$\Rightarrow$En.
LLaMA2-13B performs significantly worse than its 7B counterpart in En$\Rightarrow$xx directions due to severe off-target issues, while LexMatcher-13B improves this performance significantly.
We also consider an ICL method DictPrompt \cite{Dic_prompt_llm} which provides dictionary translations for each source word\footnote{They use Bloom-176B as the backbone and we re-implement the method on LLaMA2-13B.}, and the result shows that using dictionary translations as hints yields notable improvements in En$\Rightarrow$xx.
In contrast, LexMatcher-13B achieves better performance and is more efficient due to a much shorter context during inference.

LexMatcher demonstrates superior performance compared to other instruction fine-tuned baselines. Specifically, LexMatcher-7B outperforms Parrot-7B and TIM-7B, which construct additional translation pairs and utilize specialized instructions. 
In the En$\Rightarrow$De translation task, LexMatcher-7B surpasses TIM-7B by more than 10 BLEU and COMET points.
Moreover, LexMatcher outperforms BigTrans and ALMA consistently across the En$\Rightarrow$xx tasks, which incorporate a large amount of data for continual pretraining.
While LexMatcher-7B still underperforms GPT-3.5\footnote{GPT-3.5-turbo-0301} and GPT-4\footnote{GPT-4-0314}, the COMET scores for LexMatcher-7B are merely lower than GPT-3.5 within 2 points, and LexMatcher-13B further narrows the gap.


\begin{figure}[!t]
\centering
\includegraphics[width=1.0\linewidth]{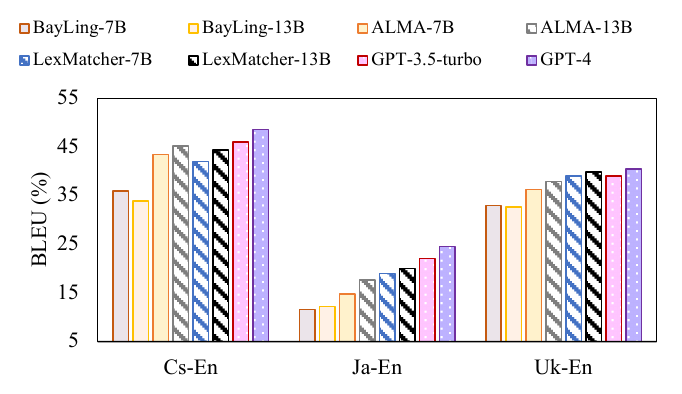}
\caption{
Zero-shot translation.
}
\label{fig:zero_shot}
\end{figure}

\paragraph{Unseen Language Directions.}
To evaluate performance in translation directions never seen previously, i.e., zero-shot multilingual capability, we further conduct experiments on
Czech-to-English (cs$\Rightarrow$en), Japanese-to-English (ja$\Rightarrow$en), and Ukrainian-to-English (uk$\Rightarrow$en).
As depicted in Figure \ref{fig:zero_shot}, 
LexMatcher-(*) exhibits superior zero-shot multilingual capability over the LLM baselines, highlighting that better aligning training languages strengthens the alignment of other languages as a by-product.

\paragraph{Disambiguation.}
By comparing the different senses of a word and multilingual expressions of meaning, the model possibly learns more precise word usage in translation.
To investigate it, we submit the models to a challenging disambiguation leaderboard, DiBiMT \cite{campolungo-etal-2022-dibimt}.
It compares the performance of NMT systems when translating sentences with ambiguous words and the performance is evaluated by accuracy.
For comparison, we display the performance of top-ranked systems including {\it DeepL}\footnote{https://www.deepl.com/en/translator}, {\it Google Translate}\footnote{https://translate.google.com}, and {\it NLLB-54B}.
The results of LLMs are from \citet{iyer-etal-2023-towards}.

\begin{table}[!t]
\centering
\small
\begin{tabular}{lccc}
\toprule
{\bf Model} & {\bf Zh} & {\bf De} & {\bf Ru} \\
\midrule
DeepL & 58.42 & \textbf{76.64} & 67.53 \\
Google-Translate & 52.09 & 67.35 & 62.03 \\
OPUS & 25.94 & 27.04 & 28.71 \\
NLLB-54B & 48.02 & 67.97 & 67.88 \\
\midrule
LLaMA-7B-ICL(1) & 30.61 & 57.41 & 60.65\\
LLaMA-7B-ICL(5) & 27.92 & 55.26 & 56.83 \\
LLaMA-65B-ICL(1) & 44.73 & 62.05 & 65.71\\
LLaMA-65B-ICL(5) & 42.49 & 62.98 & 66.31 \\
Alpaca-7B & 29.63 & 51.52 & 55.23 \\
\midrule
LexMatcher-7B & 53.28 & 63.32 & 67.72 \\
LexMatcher-13B & \textbf{59.09} & 66.98 & \textbf{69.93} \\
\bottomrule
\end{tabular}
\caption{
Accuracies on the DiBiMT benchmark which is dedicated for evaluating word disambiguation in MT.
The number following ICL denotes the number of translation demonstrations.
}
\label{exp_dis}
\end{table}

The result is shown in Table \ref{exp_dis}.
For the LLaMA models, increasing model size improves the performance, and {\it LLaMA-65B} matches {\it Google Tranlate} and {\it NLLB-54B} with few-shot prompting. 
{\it Alpaca-7B} works well without demonstration (i.e., zero-shot prompting) and significantly outperforms the supervised NMT system OPUS, which indicates its potential for further improvement through fine-tuning on translation data.
{\it LexMatcher-7B} significantly outperforms {\it Alpaca-7B} and surpasses {\it Google Translate} in Chinese and Russian disambiguation. 
With a scale of 13B, it also outperforms the best {\it DEEPL} system in Chinese and Russian, achieving accuracy rates of 59.09\% and 69.93\%, respectively.
This result demonstrates the advantage of our data construction principle.

\paragraph{Terminology.}

\begin{table}[t]
\centering
\small
\setlength{\tabcolsep}{0.8mm}{
\begin{tabular}{lcccc}
\toprule
\multirow{2}{*}{\bf Model} & \multicolumn{2}{c}{\bf Zh$\Rightarrow$En} & \multicolumn{2}{c}{\bf De$\Rightarrow$En} \\
& ChrF/COMET & Suc & ChrF/COMET & Suc \\ 
\midrule
Lingua Custodia & 32.6/60.9 & 74.7 & 61.8/73.5 & 62.2 \\
VARCO & 40.5/71.5 & 80.0 & - & - \\
$\text{UEDIN}_{LLM}$ & \textbf{41.2}/\textbf{75.7} & 75.3 & 60.0/81.3 & 58.8 \\
\midrule
LexMatcher-7B & 38.2/73.2 & 84.5 & 64.3/81.9 & 80.8 \\
LexMatcher-13B & 39.1/73.6 & \textbf{85.6} & \textbf{64.5}/\textbf{82.0} & \textbf{81.5} \\
\bottomrule
\end{tabular}}
\caption{
Performance on WMT23 terminology translation test sets.
``Suc'' indicates Terminology Success Rate.
}
\label{exp_term}
\end{table}

During training, we introduce special instructions to train the model to use the provided segment pairs. 
In this experiment, we evaluate the effectiveness of the instructions on a terminology translation test set from WMT23\footnote{https://wmt-terminology-task.github.io/}. 
The numbers of sentences on Zh$\Rightarrow$En and De$\Rightarrow$En are 2640 and 2963, respectively.
The average numbers of terms per segment on Zh$\Rightarrow$En and De$\Rightarrow$En are 3.8 and 1.1, respectively.
The result is shown in Table \ref{exp_term}, and
we only present the systems achieving the best performance on a specific metric \cite{semenov-etal-2023-findings}.
{\it Lingua Custodia} and {\it VARCO} are specialized Transformer architectures to ensure the appearance of given terminology in the translation, and {\it $\text{UEDIN}_{\rm LLM}$} uses ChatGPT with terminology translation prompts.
Compared to them, our models achieve significantly higher terminology success rates, indicating a superior ability to accurately respond to the given domain-specific terminology.
On the quality metrics, our models are inferior to {\it $\text{UEDIN}_{\rm LLM}$} on Zh$\Rightarrow$En, and achieve the best results on De$\Rightarrow$En.

\section{Analysis}

\begin{figure}[!t]
\centering
\includegraphics[width=1.0\linewidth]{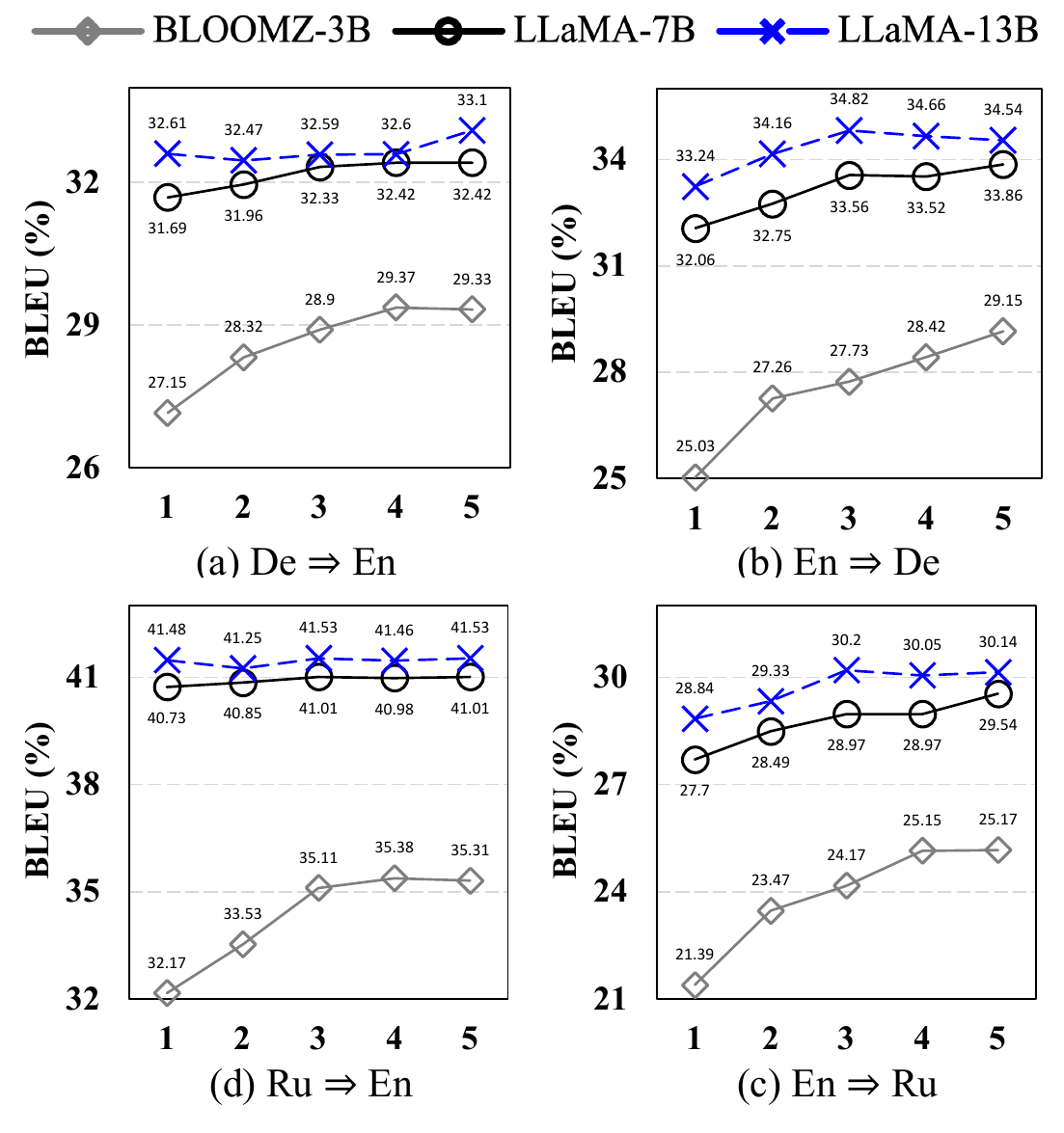}
\caption{
BLEU and COMET on the WMT22 test sets with varying $K$ and model sizes.
}
\label{fig:k_shot}
\end{figure}

\subsection{Effect of $K$}

The maximal number of bilingual contexts of each matched sense is influenced by $K$.
We show the performance of varying $K$s across different model sizes on the
WMT22 test sets (Figure \ref{fig:k_shot}).
Regardless of the amount of training data used, the larger models perform better and require less data for fine-tuning. 
In addition, the model's performance improves as $K$ increases from 1 to 3. 
With the addition of more parallel data, the performance gains begin to plateau or even slightly decrease, which aligns with the conclusions of the previous study \cite{alma}.
Thanks to the strong few-shot learning capability of the backbones, we do not need to provide as many training examples as before when training the NMT model.

\subsection{Alternative Data Selection Strategies}
\begin{figure}[!t]
\centering
\includegraphics[width=0.82\linewidth]{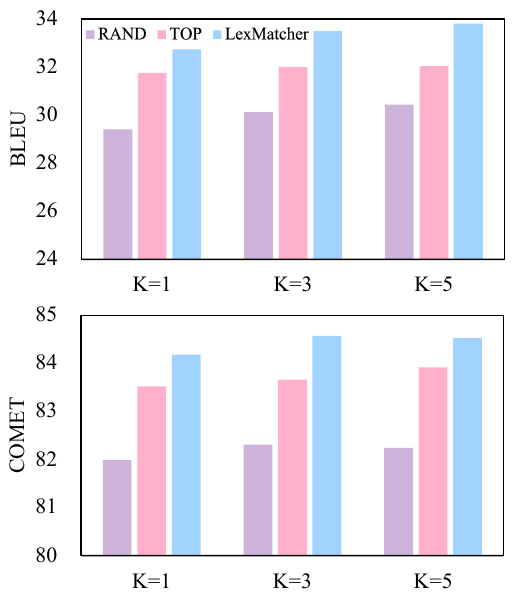}
\caption{
Performance of different data selection strategies.
}
\label{fig:select_strategy}
\end{figure}

In this experiment, we investigate two intuitive data collection methods:
1) random selection ({\it RAND}), in which the training data are randomly sampled from the corpus; and 2) quality-based selection ({\it TOP}), in which the training samples are selected based on the COMET-KIWI scores in descending order.
Specifically, we use these two methods to extract the same sample quantity as LexMatcher to mitigate the impact of sample quantity.
We use LLaMA2-7B as the backbone, and the result on WMT test sets is shown in Figure \ref{fig:select_strategy}.
The performance of {\it RAND} is inferior to the other two methods.
Random selection ensures a certain degree of diversity but the performance is uncontrollable and non-reproducible. 
{\it TOP} performs better than {\it RAND}, 
demonstrating the importance of data quality for instruction tuning.
{\it LexMatcher} can simultaneously consider both quality and diversity and achieve the best performance.

\begin{figure}[!t]
\centering
\includegraphics[width=1.0\linewidth]{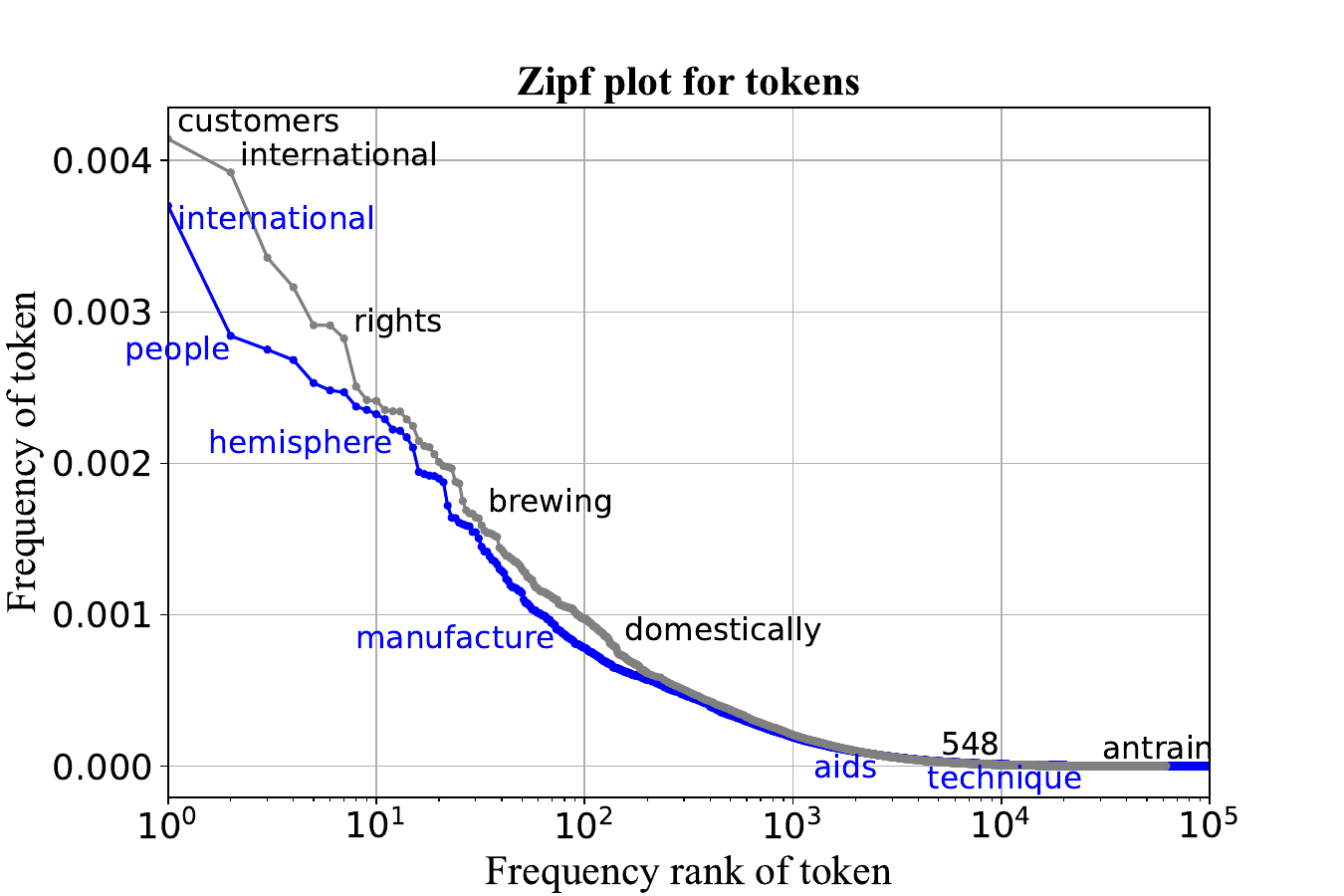}
\caption{
\label{fig:zipf}
Word frequency distributions.
The blue and gray curves denote the distributions calculated on the data selected by \cbleu{LexMatcher (K=1)} and randomly selected data, respectively. 
}
\end{figure}

\paragraph{Word Frequency Distribution}
We are interested in whether the collected data has a different word frequency distribution from the general (randomly selected) one.
We use the English data of the EN$\Rightarrow$ZH translation task with $K$=1, and plot the word frequency distributions of the collected data (blue curve) and the corresponding random data (gray curve).
As shown in Figure \ref{fig:zipf}, the blue curve tends to be smoother than the gray one, and the blue curve has more flat segments.
For words with higher frequency rankings, the word frequency of the data selected based on the dictionary is lower than that of the random data. 
This phenomenon indicates that the dictionary-based method has generated a less skewed data distribution, which could be the reason for better fine-tuning performance.
Additionally, the dictionary-based data contains 98k unique words while the random data only includes 62k unique words, indicating that the dictionary-based data covers more semantic units, thus diluting the word frequency.

\begin{table}[!t]
\centering
\small
\setlength{\tabcolsep}{0.5mm}{
\begin{tabular}{lccc}
\toprule
\multirow{2}{*}{\bf Model} & {\bf xx$\Rightarrow$En} & {\bf En$\Rightarrow$xx} &\multirow{2}{*}{\bf DiBi-Acc}\\
& {BLEU}/{COMET} & {BLEU}/{COMET} & \\
\midrule
Dev & 29.77/82.05 & 29.41/84.63 & 55.51 \\
+Supplement & 30.39/82.22 & 30.10/84.55 & 55.96 \\
+Retrieval & 32.86/82.71 & 34.13/86.27 & 59.98 \\
LexMatcher(3) & 32.71/82.61 & 34.29/86.55 & 61.44 \\
\bottomrule
\end{tabular}}
\caption{
\label{tab_aba_ave}
Ablation study on different data subsets.
}
\end{table}

\subsection{Ablation Study}
The ablation experiment of different data subsets is presented in Table \ref{tab_aba_ave}.
We use LLaMA2-7B as the backbone.
Based on the development data, simply incorporating the small amount of synthesized data generated during the data augmentation phase does not have a significant impact on the performance. 
This is possible because the data is predominantly focused on low-frequency senses, and the model is unable to effectively leverage this knowledge.
In comparison, adding the retrieved data leads to a significant performance improvement, and further introducing the synthesized data helps the model learn word disambiguation better, increasing the disambiguation accuracy from 59.98 to 61.44. 

\subsection{Combination with Other LLMs}

\begin{table}[t]
\centering
\small
\setlength{\tabcolsep}{0.5mm}{
\begin{tabular}{lcc}
\toprule
\multirow{2}{*}{\bf Model} & {\bf xx$\Rightarrow$En} & {\bf En$\Rightarrow$xx} \\
& {BLEU}/{COMET} & {BLEU}/{COMET} \\
\midrule
ALMA & 30.64/82.84 & 31.29/85.93 \\
+LexMatcher(1) & 32.34/83.11 & 33.50/86.42 \\
+LexMatcher(2) & 31.88/83.07 & 33.31/86.47 \\
+LexMatcher(3) & 33.37/83.32 & 35.30/87.09 \\
LLaMA3-8B \\
+LexMatcher(1) & 33.15/83.26 & 34.20/86.58 \\
+LexMatcher(2) & 33.29/83.26 & 35.12/87.00 \\
+LexMatcher(3) & 33.74/83.29 & 35.38/86.97 \\
Gemma-2B \\
+LexMatcher(1) & 31.68/82.42 & 31.01/84.83 \\
+LexMatcher(2) & 31.83/82.39 & 32.13/85.50 \\
+LexMatcher(3) & 31.93/82.43 & 32.33/85.66 \\
\bottomrule
\end{tabular}}
\caption{
\label{tab_comb_alma}
The performance of LexMatcher combined with different LLMs.
}
\end{table}

In this section, we investigate the performance of our data curation on different LLMs including ALMA-7B \cite{alma}, LLaMA3-8B, and Gemma-2B \cite{gemma}, and the results are shown in Table \ref{tab_comb_alma}.
ALMA \cite{alma} is the post-trained LLaMA2 on a large amount of monolingual data mixed by different languages.
We find that adding the parallel sentences constructed by LexMatcher further enhance its performance, indicating the compatibility of monolingual continual pretraining and supervised fine-tuning. 
Although the use of monolingual data during pretraining can reduce the dependency on bilingual data, the direct application of bilingual data for fine-tuning can be more resource-efficient. 
The size of parallel data collected by LexMatcher is considerably smaller than that of mixed monolingual data, and the training process is only a single stage.
Furthermore, 

\subsection{Compositional Generalization}

\begin{table}[t]
\centering
\small
\begin{tabular}{lccc}
\toprule
{\bf Model} & BLEU & Instance & Aggregate \\
\midrule
Transformer & 59.5 & 28.4 & 62.9 \\
Transformer+CReg & 61.3 & 20.2 & 48.3 \\
\midrule
LLaMA2-ICL & 38.9 & 68.6 & 87.4 \\
LLaMA2-SFT & 62.4 & 18.5 & 43.9 \\
LexMatcher & 63.5 & 15.6 & 37.3 \\
\bottomrule
\end{tabular}
\caption{
Compound translation error rates (CTERs) on CoGnition.
Instance and Aggregate denote the instance-level and aggregate-level CTERs, respectively.
}
\label{cgmtexp}
\end{table}

We investiage the effect of a more balanced atom distribution on CoGnition \cite{li-etal-2021-compositional}.
The evaluation metrics include instance-level CTER which denotes the translation accuracy of novel compounds, and aggregate-level CTER which measures the translation consistency across different contexts.
We use the data retrieval of LexMatcher to obtain 70,272 parallel sentences from the full training data (196,246) with $K$=50.
For LLM, we apply ICL with 8 examples and fine-tune LLaMA2-7B on the randomly sampled training data, of which the size is similar to the retrieved data.
The results are shown in Table \ref{cgmtexp}.
ICL does not yield good compositional generalization performance, while the fine-tuned LLaMA2 outperforms the previous NMT models significantly. 
{\it LexMatcher} achieves lower compound translation error rates than SFT with the same amount of training data, demonstrating the positive effect of the more balanced data distribution.

\section{Conclusion}

In this paper, we presented LexMatcher, a dictionary-centric data curation method for supervised fine-tuning smaller-sized LLMs to better translation models.
We use the bilingual dictionary as the pivot and try to collect limited parallel sentence pairs to cover the senses uniformly.
Experiments and analyses validate the effectiveness of LexMatcher from multiple perspectives including zero-shot translation, disambiguation, and terminology translation.
One potential avenue for future research involves extending LexMatcher to low-resource scenarios, where the utilization of monolingual data is crucial for achieving satisfactory translation performance.

\section{Limitations}
This work focuses solely on improving translation performance for medium and high-resource language pairs. 
For low-resource language pairs that inherently lack parallel data, it is crucial to explore how to optimize LLMs on such translation tasks by integrating dictionaries, monolingual, and possible bilingual data.

\bibliography{anthology,custom}

\appendix

\begin{figure*}[!t]
\centering
\includegraphics[width=0.9\linewidth]{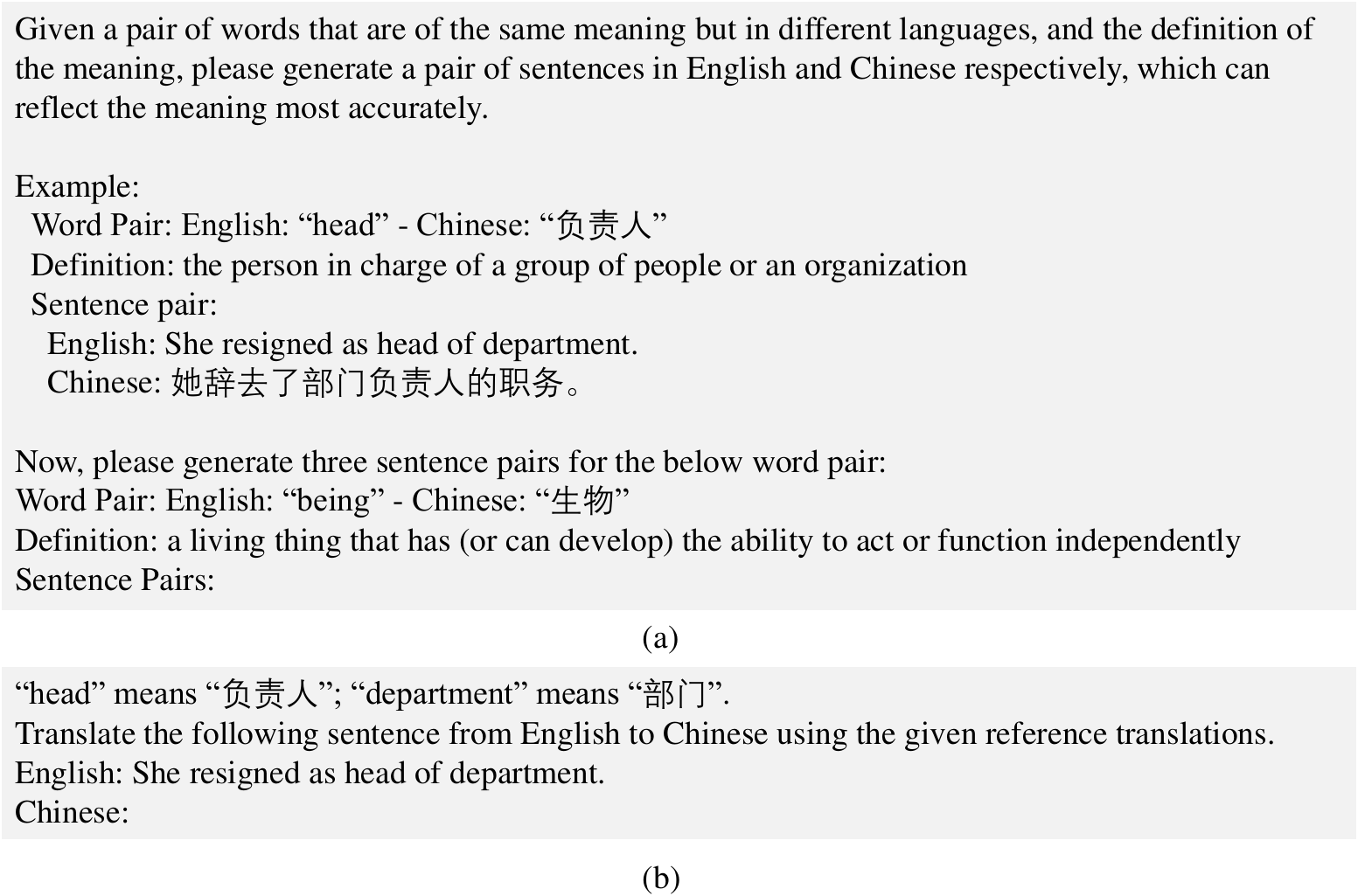}
\caption{
Prompts used for (a) manipulating ChatGPT to generate translation demonstrations and (b) terminology translation.
}
\label{fig:prompt_gpt}
\end{figure*}

\begin{table*}[!t]
\centering
\small
\renewcommand{\arraystretch}{1.1}
\begin{spacing}{1.1}
\resizebox{\textwidth}{!}{
\begin{tabular}{lcccccc}
\toprule
\multirow{2}{*}{\bf Model} &
 \multicolumn{1}{c}{\bf Zh$\Rightarrow$En} & \multicolumn{1}{c}{\bf En$\Rightarrow$Zh} & \multicolumn{1}{c}{\bf De$\Rightarrow$En} & \multicolumn{1}{c}{\bf En$\Rightarrow$De} &
 \multicolumn{1}{c}{\bf Ru$\Rightarrow$En} & \multicolumn{1}{c}{\bf En$\Rightarrow$Ru} \\
& {BLEU}/{COMET} & {BLEU}/{COMET} & {BLEU}/{COMET} & {BLEU}/{COMET} & {BLEU}/{COMET} & {BLEU}/{COMET} \\
\midrule
Dev & 23.59/78.94 & 35.43/84.28 & 29.04/83.63	& 28.58/84.09 & 36.68/83.58	& 24.23/85.54 \\
+Supplement & 23.69/79.05 & 36.50/84.20	& 29.45/83.82	& 28.67/83.98 & 38.03/83.80 & 25.14/85.49 \\
+Retrieval & 25.36/79.46 & 40.14/86.01 & 32.37/84.31 & 33.26/85.77 & 40.86/84.36 & 29.00/87.03 \\
LexMatcher(3) & 24.81/79.13 & 40.34/86.11	& 32.33/84.29 & 33.56/86.31 & 41.01/84.43 & 28.97/87.23 \\
\midrule
ALMA-7B \\
+LexMatcher(1) & 24.27/79.82 & 31.77/84.52	& 41.00/85.01 & 38.61/85.83	& 33.12/86.19 & 28.77/87.25 \\
+LexMatcher(2) & 24.04/79.88	& 38.27/85.93 & 31.39/84.32 & 32.85/86.14 & 40.61/85.07 & 28.82/87.34 \\
+LexMatcher(3) & 25.20/80.21 & 41.40/86.59 & 32.49/84.49 & 34.44/86.66 & 42.42/85.28 & 30.07/88.02 \\
LLaMA3-8B \\
+LexMatcher(1) & 26.40/80.47 & 40.30/86.11 & 32.44/84.52 & 33.16/86.09 & 40.63/84.79 & 29.15/87.54 \\
+LexMatcher(2) & 26.33/80.31 & 42.34/86.94 & 32.36/84.54 & 33.68/86.37 & 41.19/84.93 & 29.36/87.69 \\
+LexMatcher(3) & 26.89/80.51 & 41.88/86.74 & 32.95/84.46 & 34.22/86.49 & 41.39/84.92 & 30.04/87.70 \\
Gemma-2B \\
+LexMatcher(1) & 24.88/79.75 & 37.89/85.01 & 31.35/83.77 & 29.27/83.95 & 38.81/83.75 & 25.87/85.53 \\
+LexMatcher(2) & 25.19/79.60 & 39.53/85.92 & 31.43/83.77 & 30.35/84.51 & 38.87/83.81 & 26.53/86.09 \\
+LexMatcher(3) & 24.84/79.55 & 39.19/85.98 & 31.77/83.80 & 30.81/85.04 & 39.18/83.95 & 27.00/85.98 \\
\bottomrule
\end{tabular}
}
\end{spacing}
\caption{
\label{tab_aba}
Detailed results of ablation study and combination with different LLMs.
}
\end{table*}

\section{Computational Details}
We conducted experiments using the Huggingface Transformers.
The experiments are performed on NVIDIA A100 GPU, and all the results are run once with the random seed 42.
According to the data license of WMT22, the data released for the General MT task can be freely used for research purposes.

\section{Prompts Used for Manipulating ChatGPT and Terminology Translation}\label{app_chatgpt}
The prompt used to manipulate ChatGPT consists of three parts (Figure \ref{fig:prompt_gpt} (a)).
The first part is used to describe the task: generate a pair of parallel sentences, which can reflect the meaning of a given segment pair accurately.
The second part is an example to demonstrate the format of the input and output including a segment pair, a definition of the sense, and a sentence pair.
The third part is the segment pair requires translation demonstration.

The prompt for terminology translation is shown in Figure \ref{fig:prompt_gpt} (b).

\section{Corpus Preprocessing}\label{app_filter}

Since the filtered data of Russian$\Leftrightarrow$English is significantly less than the other language pairs, we introduce the training set from Tatoeba translation challenge 2021\footnote{https://github.com/Helsinki-NLP/Tatoeba-Challenge/tree/v2021-08-07/data}.
We filter data with the commonly used rule-based methods
and model-based QE.
The rules include the following categories: 
(1) sentence-level deduplication, 
(2) filter out the sentences longer than 100 words or contain a single word exceeding 40 characters,
(3) remove sentence pairs where the ratio of source sentence length to target sentence length is significantly different, i.e., below 1/3 or above 3,
(4) filter out the sentences with high repeat ratio, i.e., the proportion of the frequency of the most frequent word in a sentence to the total word frequency greater than 0.3, 
and 
(5) filter out the sentences in which the proportion of the content words is between 0.3 and 0.8.
In this way, low-quality data can be efficiently filtered out, saving time and resources for the subsequent model-based QE.

We utilize one of the state-of-the-art QE models, COMET-KIWI\footnote{https://huggingface.co/Unbabel/wmt22-cometkiwi-da}, to obtain sentence-level quality scores.
For every sentence pair in the training data, we calculate the QE score for the translation from English to the foreign language. 
These scores are utilized for both translation directions, as evaluating both directions of the training data can be computationally expensive.
We remove sentence pairs with low data quality, e.g., those that have a score below 40.
We use spaCy\footnote{https://spacy.io/} for the lemmatization.




\end{CJK}
\end{document}